\pdfoutput=1

\documentclass[11pt]{article}

\usepackage{acl}

\usepackage{times}
\usepackage{latexsym}

\usepackage[T1]{fontenc}

\usepackage[utf8]{inputenc}

\usepackage{microtype}

\usepackage{float}
\usepackage{url}
\usepackage{bm}
\usepackage{amssymb}
\usepackage{dsfont}
\usepackage{multirow}
\usepackage{comment}
\usepackage{multicol}
\usepackage{multirow}
\usepackage{booktabs}
\usepackage{xspace}
\usepackage{cleveref}
\usepackage{tabularx}
\usepackage[linewidth=1pt]{mdframed}
\usepackage{graphicx}
\usepackage{subcaption}
\usepackage[graphicx]{realboxes}
\usepackage{algorithm}
\usepackage{algpseudocode}

\newcolumntype{P}[1]{>{\centering\arraybackslash}p{#1}}

\definecolor{c1}{HTML}{4e79a7}%
\definecolor{c2}{HTML}{f28e2b}%
\definecolor{c3}{HTML}{009E73}%
\definecolor{c4}{HTML}{56B4E9}%
\definecolor{c5}{HTML}{CC79A7}%
\definecolor{c6}{HTML}{E69F00}%
\definecolor{c7}{HTML}{844E4D}%
\definecolor{c8}{HTML}{2D512A}%

\definecolor{oorange}{HTML}{d95f02}
\definecolor{bblue}{HTML}{7570b3}
\definecolor{ggreen}{HTML}{1b9e77}
\definecolor{ppurple}{HTML}{e37fbb}
\definecolor{lgreen}{HTML}{9CD24A}
\definecolor{yyellow}{HTML}{FFD52D}
\definecolor{ggold}{HTML}{E1BC89}
\definecolor{ggray}{HTML}{AAAAAA}

\usepackage{listings}
\definecolor{dkgreen}{rgb}{0,0.6,0}
\definecolor{gray}{rgb}{0.5,0.5,0.5}
\definecolor{mauve}{rgb}{0.58,0,0.82}
\newcommand{\ourdataset}{\textsc{Tailor-CGO}\xspace}

\usepackage{tikz}

\title{Tailoring Vaccine Messaging with Common-Ground Opinions}

\author{\\ \bf
    Rickard Stureborg \hspace{4mm}
    Sanxing Chen \hspace{4mm}
    Ruoyu Xie \hspace{4mm}
    Aayushi Patel \hspace{4mm}
    Christopher Li \AND 
    \\ [-2.5ex] \bf 
    Chloe Qinyu Zhu \hspace{4mm}
    Tingnan Hu \hspace{4mm}
    Jun Yang \hspace{4mm} 
    Bhuwan Dhingra \\ [1ex]
    Duke University \\ [1ex]
    \texttt{\{rickard.stureborg, sc829, rx55,}\\
    \texttt{akp43, cl619, qz124, th311\}@duke.edu},\\ 
    \texttt{\{junyang, bdhingra\}@cs.duke.edu} 
    \\
    }

\begin{document}
\maketitle
\begin{abstract}

One way to personalize chatbot interactions is by establishing common ground with the intended reader.
A domain where establishing mutual understanding could be particularly impactful is vaccine concerns and misinformation.
Vaccine interventions are forms of messaging which aim to answer concerns expressed about vaccination.
Tailoring responses in this domain is difficult, since opinions often have seemingly little ideological overlap.
We define the task of tailoring vaccine interventions to a Common-Ground Opinion (CGO).
Tailoring responses to a CGO involves meaningfully improving the answer by relating it to an opinion or belief the reader holds.
In this paper we introduce \ourdataset, a dataset for evaluating how well responses are tailored to provided CGOs.
We benchmark several major LLMs on this task; finding GPT-4-Turbo performs significantly better than others.
We also build automatic evaluation metrics, including an efficient and accurate BERT model that outperforms finetuned LLMs,
    investigate how to successfully tailor vaccine messaging to CGOs,
    and provide actionable recommendations from this investigation.\footnote{
\ourdataset dataset and code available at: \url{https://github.com/rickardstureborg/tailor-cgo}}

\end{abstract}

\section{Introduction}

\begin{figure}[t]
   \centering
   \includegraphics[width=.48\textwidth]{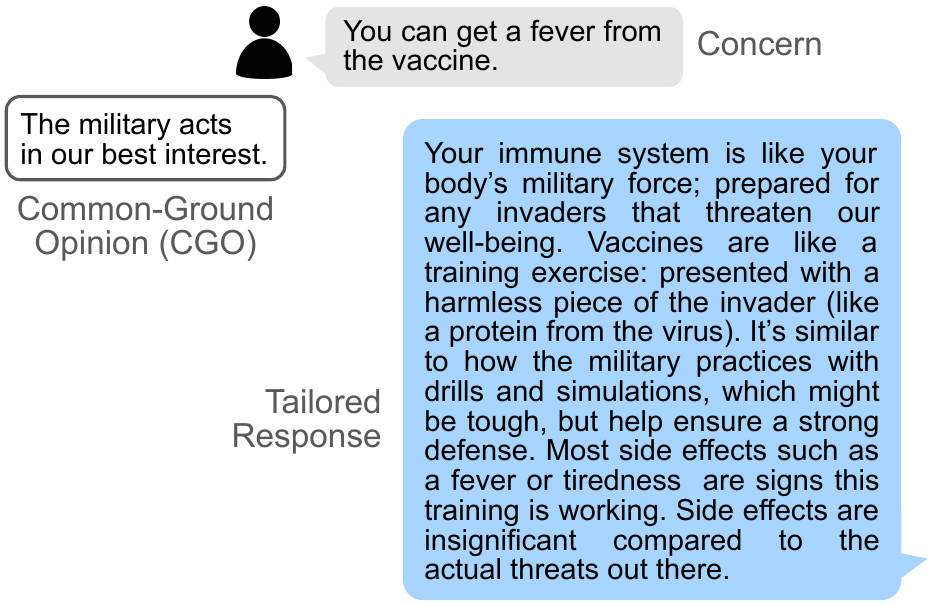}
   \setlength{\abovecaptionskip}{-10pt}
   \setlength{\belowcaptionskip}{-2pt}
   \caption{
        \textbf{Example of a Tailored Response to answer a Vaccine Concern while Tailoring to a Common-Ground Opinion}. 
        The response above is a shortened version of a model response for the task of tailoring to CGOs.
        The responses is able to relate two seemingly unrelated topics: side effects from vaccination and a strong support for the military.
        It is strengthened by language and analogies that may appeal to the user without becoming manipulative.
        This work creates an evaluation framework and benchmarks different LLMs on their ability to generate such tailored responses.
    }
    \label{fig:short_example}
\end{figure}

Finding common-ground has long been understood as an important aspect of language and communication \citep{gibbs1988commonground}.
Creating this connection during communication can establish a sense of trust and is an inclusive goal to strive for when preparing personalized messaging.
We can establish common ground on many things, such as beliefs or opinions, although it may be unclear whether such opinions can be seamlessly related to the main message being communicated.

One particularly relevant use case for establishing common ground is vaccine messaging.
Loss of trust and polarization are increasingly pervasive issues impacting vaccine hesitancy \citep{ozawa2013trust}.
Those involved with vaccine messaging (e.g. vaccine navigators, public health organizations, healthcare professionals) could be aided by further tools to help 
    write responses, 
    personalize information,
    or generate creative ideas for their messaging efforts.

Large Language Models (LLMs) have demonstrated high performance in controlled text generation.
Recent research has increasingly explored personalization given the abilities of these models \citep{soni-etal-2022-human, welch-etal-2022-leveraging, salemi2023lamp} with many pointing out remaining challenges \citep{kirk2023personalisation}.
How LLMs handle personalization in terms of opinions and beliefs remains under-explored.

To this end, we propose a task for tailoring vaccine messaging towards common-ground opinions (\autoref{fig:short_example}).
By providing common-ground opinions and instructing LLMs to tailor towards them, we hope to help address the imbalance and inequity in the broader information landscape surrounding vaccination.
Given that some work \citep{santurkar_2023_opinions} has pointed out these models may exhibit biased opinions that do not reflect diverse or underrepresented groups, we highlight and evaluate current major LLMs' ability to perform this task by building comprehensive automatic evaluation metrics.

This paper provides the following contributions:
\begin{itemize}
  \setlength{\itemsep}{0pt}
  \setlength{\parskip}{0pt}
  \setlength{\parsep}{0pt}
    \item Evaluation of several major LLMs on their ability to use `Common-Ground Opinions' in vaccine messaging.
    \item \ourdataset, a comprehensive and high-quality dataset for training and evaluation.
    \item Actionable recommendations of which CGOs are most useful to address a given concern.
\end{itemize}

\section{Related Work}

Personalization has been explored in various NLP tasks, most notably within the domain of dialogue response generation~\citep{wang2019persuasion,zhang-etal-2018-personalizing,zheng2019personalized,joshi2017personalization}.
Using natural language prompts, language models can generate texts that align with demographics or identities, reflecting cross-cultural values~\citep{arora2023probing}, political ideology~\citep{simmons2023moral}, or opinions on societal issues~\citep{argyle2023lmhuman, santurkar_2023_opinions}, or infer personal attributes~\citep{wang-etal-2022-extracting}.
Previous work has also investigated balancing benefits and harms of personalization~\citep{kirk2023personalisation}, providing benchmarks~\citep{salemi2023lamp}, and constructing user-conditioned language models~\citep{soni-etal-2022-human, welch-etal-2022-leveraging}.
However, we are unaware of any work to date which examines personalization of LLM responses within the vaccine misinformation domain, or work which focuses on tailoring messages to common-ground opinions.
There is also a line of work in NLP on grounding responses~\citep{cho2020grounding, chandu-etal-2021-grounding, zhou-etal-2022-reflect}.

Various benchmarks have been proposed to test general LLM abilities in generating task-specific responses~\citep{hendrycks2021measuring,khashabi-etal-2022-genie,zheng2023judging}.
Our benchmark emphasizes on a coverage of diverse opinions for controllable generation in the domain of vaccine concern and misinformation.
In the era of LLMs, we have seen a renewed interest in automatic evaluation metrics of text generation, due to the need for reinforcement signals~\citep{stiennon2020learning,rafailov2023direct}.
We demonstrate the feasibility to build automatic evaluation metrics using our data, facilitating future efforts to improve LLM generation.

Previous research of misinformation has explored classification of common concerns and misinformation topics \citep{coan_climate_2021, stureborg2023vaxconcerns, Zhu2024HierarchicalMC}, fact-checking statements \citep{thorne_fever_2018}, or claim review \citep{arslan_benchmark_2020, arslan_modeling_nodate}, which determines if claims are worth fact-checking.
While addressing these concerns and misinformation is important, our work aims to begin addressing the vaccine misinformation through tailored messaging.

Indeed, there is already substantial work on establishing common ground for the goal of successful communication. It is well understood in cognitive sciences~\citep{clark1982, clark1989, clark1991}.
Likewise, arguments tailored to information about the target audience have been shown more effective~\citep{hirsh_personalized,hadoux_comfort}.
However, reliably generating such responses automatically is an open research question which could serve as the foundation for future research into effective communication practices or persuasion.

\section{\ourdataset Dataset Creation}
\label{sec:dataset}

In this section we describe the task of tailoring a response to a CGO, introduce and describe the dataset and its components, and outline how the dataset is labeled.
The final dataset contains $22,400$ unique tailored responses from 6 different LLMs, labeled with a mix of absolute scores or pairwise rankings.

\subsection{Task Definition}
\label{task_definition}
We define our task of tailoring to common-ground opinions as follows.
In response to an expressed \emph{concern} about vaccination,
the task is to generate an intervention tailored to a given \emph{common-ground opinion (CGO)},
which should act as the basis for framing the response.

A successfully tailored response should meet five criteria: 
(1) It should fully answer the concern to promote vaccination or encourage engaging further with health professionals. 
(2) The opinion should be used or referred to in the response, either directly or indirectly. 
(3) The response should accept the opinion as true, rather than refute it. 
(4) The answer to the concern should be meaningfully linked to the opinion in some manner.
Finally, (5) the use of the opinion should strengthen the response to the expressed concern, such that the removal of the opinion would weaken the response.

\subsection{Concerns and Opinions Statements}
\label{sec:concerns_and_opinions}
To provide inputs for our task, we need explicitly stated concerns and opinions.
For concerns, we utilize the VaxConcerns taxonomy from \citet{stureborg2023vaxconcerns} and then prompt
GPT-4 to generate a large variety of concern statements,
as detailed in \Cref{apx:concern_statements}.
Some of these statements refer to specific vaccines (COVID-19, HPV, MMR, Influenza, and Yellow Fever), while others are agnostic towards vaccine type.
There are 1166 total concern statements, all cleaned by the authors of this paper and mapped to the VaxConcerns taxonomy.
One example concern is shown below, where the sampled vaccine-type was \textit{`MMR'} and the concern category was \textit{`2.4: Lack of benefits $\to$ Insufficient risk'}.
\begin{quote}
    ``Measles, mumps, and rubella cases are so rare nowadays, the MMR vaccine seems unnecessary.''
\end{quote}

For opinions, we generate statements by paraphrasing questions from OpinionQA \citep{santurkar_2023_opinions}.
These public opinion survey questions (originally sourced from PEW Research polls) have awkward phrasing which make them difficult to use as-is.
Therefore, we use GPT-4 to convert these questions to single-sentence statements expressing the opinion as a fact.
For example, one such opinion statement reads:
\begin{quote}
    ``In general, society tends to look up to men who are manly or masculine these days.''
\end{quote}

\subsection{Response Generation}
\label{sec:response_generation}
To produce candidate responses for annotation, we prompt LLMs to tailor to common-ground opinions.
However, we do not want downstream uses of our dataset to rely heavily on a specific combination of model and prompt.
Therefore, we use a variety of systems for producing candidate responses.
We conducted an extensive qualitative analysis of the generated responses to find common issues and strengths, which are detailed in \Cref{apx:qualitative_analysis}.

We generate a mix of candidate responses that allow for both
    \textbf{intra-opinion} comparisons, where responses are tailored to the same opinion, and 
    \textbf{inter-opinion} comparisons, where responses are tailored to different opinions. 
Specifically, we create ``blocks'' of 4 response generations, with each block structured as follows:
\begin{lstlisting}
    Concern1 + Opinion1 -> ResponseA    (rA)
    Concern1 + Opinion1 -> ResponseB    (rB)
    Concern2 + Opinion2 -> ResponseC    (rC)
    Concern2 + Opinion3 -> ResponseD    (rD)
\end{lstlisting}
Here, $(rA, rB)$ provides an intra-opinion comparison and $(rC, rD)$ provides an inter-opinion comparison.
For each block, we randomly sample 2 concerns and 3 opinions with replacement.
For response generation, we randomly sample model, prompt, and temperature from their respective domains as explained in the following subsections.
For half of our blocks, we fix these system parameters (model, prompt, temperature) between $rA, rB$ and between $rC, rD$.
This design gives opportunities to compare responses sampled under identical settings as well as different ones.

We manually inspected 700 sample responses of the 1546 unguided responses.
Within this sample, we found 21 instances of causes of clear response failure, demonstrating a 3\% failure rate.
There were several distinct failure modes that we identified with varying frequencies:
    (a) 1.9\% - The model directly or indirectly assumes the identity of a human 13 times,
    (b) 0.3\% - The model assumes an identity for the person it is responding to 2 times,
    (d) 0.3\% - The model responds to a different vaccine concern than that which is provided in the prompt 2 times,
    (f) 0.3\% - The model crafts a response that contains a template element like “hey [friend’s name]” 2 times
    (c) 0.1\% - The model explicitly disagrees with the CGO 1 time, and 
    (e) 0.1\% - The model makes a factually incorrect or ambiguous statement 1 time.

\paragraph{Models}
We consider six models: Llama-2, Vicuna, WizardLM, GPT-3.5, GPT-4 and GPT-4-Turbo.
In early experiments, such as collection of the dev set, we use smaller model sizes ($13$b models) for candidate response generation.
However, in the final round-2 and round-3 training and test data, we use the most powerful models possible on our hardware (70B parameter models).
A full list of the model checkpoints we use, with complete citations and links, are shown in \Cref{tab:models_used}.
We randomly sample temperature uniformly between 0 and 1 during generation to encourage a diversity of both creative and high-likelihood outputs.

\begin{table}[h]
    \caption{\textbf{Models checkpoints used in this work}. 
    }
        \centering
        \small
        \begin{tabular}{l}
        \toprule
        Model Checkpoint \\
        \midrule
        llama-2-13b-chat-hf\\
        llama-2-70b-chat-hf\\
        vicuna-13b-v1.5\\
        vicuna-33b-v1.3\\
        WizardLM-13B-V1.2\\
        WizardLM-70B-V1.0\\
        gpt-4-1106-preview\\
        gpt-4-0613\\
        gpt-3.5-turbo-0613\\
        \bottomrule
    \end{tabular}
    \label{tab:models_used}
\end{table}

\paragraph{Prompting}
\label{sec:prompting_method}
To increase the diversity of our prompts, we make use of role-playing \citep{wang2023rolellm}, chain-of-thought \citep{wei2022cot}, and pointed instructions for generation \citep{ouyang_instruct}.
We consider all combinations of the following strategies:
\begin{itemize}
    \item \textbf{Roles}: We ask the model to produce different text styles by instructing it with a role (e.g. parent, doctor, redditor) to take when responding.
    We use 10 different roles.
    \item \textbf{Chain of Thought} (CoT): The model is asked to first think through the process of generating the tailored response by writing out a plan, and is then subsequently prompted to produce the final response.
    We compare the quality of responses with and without CoT.
    \item \textbf{Guidelines}: The prompt includes written instructions for principles to follow or general behaviors to avoid. 
    These are sourced from 
        1) CDC guidelines on vaccine messaging, 
        2) the criteria (\S\ref{task_definition}) for what makes well-tailored responses, and 
        3) general guidelines to avoid issues noticed in responses during the development phase.
    To avoid increasing the size of the prompt prohibitively, and to ensure diversity, we randomly sample five such guidelines from all sources uniformly and include them in prompts marked with the `guided' flag.
    Approximately half of responses are produced with a guided prompt.
\end{itemize}
Overall, these strategies are each randomly sampled, leading to 40 potential prompts types, with more than 500K possible unique combinations (due to the random sampling of guidelines).\footnote{
    40 prompt types come from the 10 roles, 2 settings for CoT (on or off), and 2 settings for guidelines (on or off).
    Accounting for all possible sampled subsets of 5 from the 22 guidelines gives us 500K unique combinations ($20 + 20 \cdot _{22}C_{5}$)
    
}

\subsection{Human Annotation}
\begin{figure}[h]
   \centering
   \includegraphics[width=.48\textwidth]{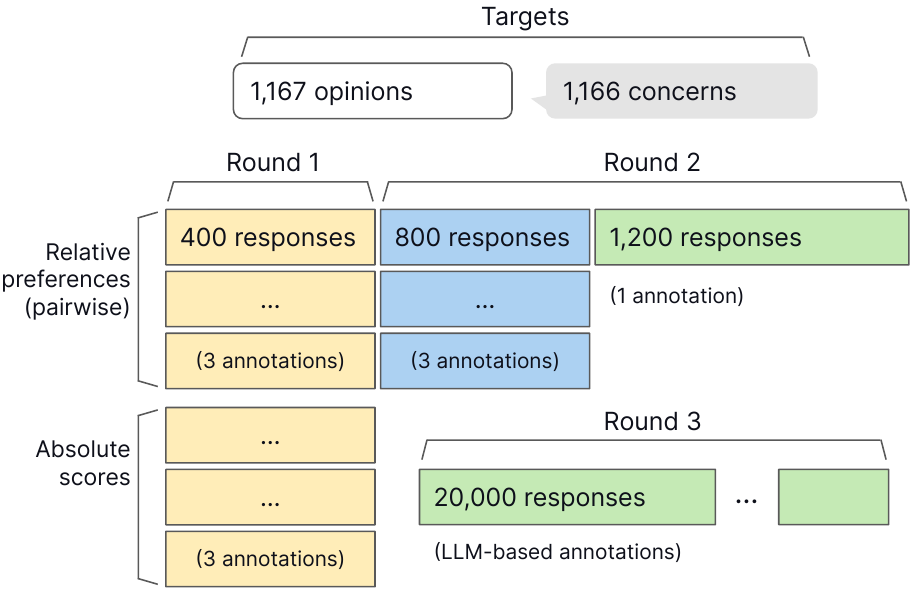}
   \setlength{\belowcaptionskip}{-15pt}
   \caption{
       \textbf{\ourdataset dataset partition sizes}.
       Colors indicate which train/dev/test split each partition is included in.
       Green = train, Yellow = dev, Blue = test.
       Relative preferences are collected by asking which of two responses is better tailored, while absolute scoring asks for a 1-5 score for a single response.
       Both Dev and Test sets (Yellow and Blue) contain 3 independently collected annotations per input response, represented by 3 stacked boxes.
       The training set (Green) contains just one annotation per response to maximize diversity.
    }
    \label{fig:dataset_partitions}
\end{figure}

We collect data in 3 rounds.
First, we annotate tailored responses using both absolute scoring and relative preferences in parallel randomly assigned conditions.
This is used as our dev set (Yellow).
We find that relative preferences yields higher data annotation quality, so in a second round we invite back the 8\% most accurate crowdsource workers to label our test (Blue) and training (Green) sets.
Third, we label a much larger set of tailored responses using LLM evaluators.
\Cref{fig:dataset_partitions} shows a useful diagram explaining the various partitions of the final dataset.
We reference these partitions frequently in this section.

\paragraph{Anotation Scheme}
To score the quality of candidate tailored responses, we collect preferences using crowdsource annotations through Amazon Mechanical Turk (AMT).
Given a particular concern-opinion and response pair, annotators are asked to judge the quality of the response based on the criteria listed under Task Definition. We consider two potential annotation schemes:

\begin{itemize}
\setlength{\itemsep}{0pt}
\setlength{\parskip}{0pt}
\setlength{\parsep}{0pt}
\item \textbf{Absolute scoring:} Annotators are asked to make absolute judgments of how well-tailored each response is on an individual basis.
These judgments are given on a 1-5 scale, ranging from \textit{Very poorly tailored} to \textit{Very well tailored}.
However, when performing within-team annotations, we found that Likert-style scales were difficult to use since it is hard to calibrate what level of quality warrants being \textit{well tailored} versus \textit{very well tailored}.
\item \textbf{Relative preference:} To circumvent the difficulty of absolution scoring, we instead ask annotators to make relative judgments, comparing two responses against one another and ranking them according to how well-tailored they are.
This label set is ordinal, with \textit{Response A} (is better), \textit{Equal}, and \textit{Response B} (is better), in order.
\end{itemize}

Our round-1 results indicated that collecting relative preferences leads to higher agreement about the eventual ranking of responses.
Annotators labeling preference of two responses directly agree on 57.3\% of labels, while marking an absolute score for a single response yielded agreement of just 44.2\%. 
One reason for this is an increase in ties: absoluting scoring yield $19.5\%$ tied preferences between pairs, while directly asking for preference in the pair results in $8.5\%$ ties.
This increase in ties is problematic not just for annotator agreement, but potentially also for labeling efficiency, since we lose information about nuanced differences in pairs with tied absolute scores.

\paragraph{Annotator Selection}
We train our crowdsource workers using a short 9-example tutorial, where they first annotate and are then given feedback with reasons motivating the correct choice as well as highlights over the responses with hover-text providing further explanations.
We open our annotation task to all annotators which have an AMT `masters' qualification and at least 2,500 approved HITs at an approval rate of at least 99\%.
These very selective criteria ensure only the highest performing annotators.
After the tutorial, annotators complete a short (3-example) entrance exam of easier, expert-labeled questions to further qualify;
we remove any worker who incorrectly answers at least one of the questions.
During annotation, we randomly insert, for 5\% of shown examples, attention checks that instruct workers to select certain options to ensure they are fully reading the passages.

We invite only the top 8\% of annotators for round-2 annotations based on their scores on the tutorial examples.
We pay approximately 15-20 USD per hour for the workers in our round-1 annotation, and 25-30 USD per hour for those in round-2 annotations. We offer \$100 bonuses to the top 25\% of annotators in round 2 to incentivize high-quality annotations, and regularly examine their annotations and offer feedback through direct messaging during data collection.

Further details of the annotation platform, including screenshots, are available in \Cref{apx:annot_platform}

\paragraph{Inter-annotator Agreement}
\label{sec:agreement}
Round-2 annotations (test set marked in blue and human-based training set marked in green in \Cref{fig:dataset_partitions}) show much higher agreement than round 1.
As shown in \Cref{tab:agreement_by_round},
crowdsource annotators agree on 73.2\% of labels, comparable to previous works \citep{stiennon_summarize_hf, ouyang_instruct} on annotating human preferences for reward modeling.

\begin{table}[t]
        \centering
        \small
        \begin{tabular}{lc}
        \toprule
        Annotation Strategy & Percent agreement \\
        \midrule
        Dev set (round 1 - absolute) & 44.2\% \\
        Dev set (round 1 - relative) & 57.3\% \\
        Test set (round 2 - relative) & 73.2\% \\
        \citet{bai2022training} & 63.0\% \\
        \citet{stiennon_summarize_hf} & 73.0\% \\
        \citet{ouyang_instruct} & \textbf{77.3}\% \\
        \bottomrule
    \end{tabular}
    \caption{\textbf{Agreement between crowdsource annotators when presented with two tailored responses}. 
    Our test set shows comparable amounts of agreement to work in instruction tuning, despite a highly subjective and difficult task and offering 3 options (\textit{A}, \textit{Equal}, \textit{B}) rather than two (\textit{A}, \textit{B}) in the annotation interface.
    Note that the human-labeled training set was collected together with the test set, but does not have multiple annotations and is therefore left out of this table.
    }
    \label{tab:agreement_by_round}
\end{table}

\section{Automatic Evaluation}
\label{sec:auto_eval}
While human annotation shows a high agreement on evaluating LLM responses,
conducting it at scale to study the various settings we consider is prohibitively expensive.
Hence, in this section we use the annotations collected above to develop automatic evaluation metrics
for the \ourdataset task.
The goal of automatic evaluation is to provide a cheap alternative for labeling candidate responses and automatically evaluate or compare models.
It also allows for deeper analysis of trends when human annotation becomes restricted by scale.
We describe our approach below.
First, we prompt a generative model (GPT-4-Turbo) to score responses directly.
Second, to further reduce cost, we fine-tune open-source language models (BERT and Llama-2) using the results from the former and/or human annotations.

\subsection{Zero-shot Prompting}
We use GPT-4-Turbo as an automatic evaluator by prompting it in a zero-shot setting.
Specifically, we use G-Eval's \citep{GEval} instruction template by replacing the definition and evaluation steps in its provided prompts with descriptions explaining what a well-tailored response is.
G-Eval is an automatic evaluation framework for text summarization built on top of GPT models, but has been adopted to other tasks as well.

Because log probabilities are no longer available for GPT models, we sample 100 predictions at temperature 1.0 for each response, stopping the model outputs at 10 tokens.
We then parse these outputs to collect discrete 1-5 scores, and a mean score is calculated over the 100 samples thereby approximating the original weighted prediction by token probabilities used in G-Eval.
We provide all outputs (raw and cleaned) in our dataset.
When needed, these absolute scores can be paired together with another response answering the same concern, and by comparing the scores a relative preference can be reported.
This is how we perform evaluation on the Test set, and also how we provide large labeled datasets to our finetuning methods.

\subsection{Fine-tuning}
We further explore the possibility of using open-source language models to perform automatic evaluation, thus reducing the cost on closed-source API inference.
We test both an encoder-only model (BERT-base) and an auto-regressive model (Llama-2-13b) in a
knowledge distillation setup, where they are trained on a large dataset of responses scored
by GPT-4-Turbo.
We continue finetuning beyond the Round 3 dataset by using the human-labeled data from the Dev set.
In the relative preference setting, this is done by either randomly mixing in the human labeled data, or training on this data after the model converges.
In the absolute scoring setting, we use a margin ranking loss.

\paragraph{BERT Fine-tuning}
We fine-tune a 110M BERT model~\cite{devlin-etal-2019-bert}.
For absolute scoring, we train for regression on a 1-5 score.
For relative comparison, we train as a binary classification task (as opposed to using a parallel contrastive loss sometimes used in such similar settings \citep{seth2023visual}).
The model takes as inputs the text of a sampled response pair from the round-3 training data.
On top of this model, we add a linear layer that maps the final BERT layer hidden states into one (absolute scoring) or two scalars (relative preferences).
Mean-squared error and cross-entropy loss are applied correspondingly.
We sample approximately 10,000 responses (pairs) along with their scores from GPT-4-Turbo to construct the training data.
We train using a batch size of 8 for 5 epochs, and the AdamW~\cite{loshchilov2018decoupled} optimizer with a learning rate of 2e-5, 10\% linear warm-up, and linear decay to 0.
Training and evaluation is done on a single NVIDIA A5000 GPU.

\paragraph{Llama-2 Fine-tuning}
For both absolute scoring and relative preference settings, we use a 13B LLaMA-2 as the base model for fine-tuning. 
We utilize QLoRA \citep{dettmers2023qlora} for computational efficiency and use the AdamW optimizer to train the model for 5 epochs with a batch size of 4. 
The training is conducted on 4 NVIDIA RTX A6000 GPUs, setting the learning rate to 2e-4 with a 3\% warmup. 
The LoRA rank and alpha are set to 64 and 16, respectively, with a 0.1 dropout between the two matrices.

We use Alpaca's training prompt format \citep{Bommasani2021FoundationModels}
    where the instruction is replaced by the evaluation instruction and metrics, 
    the input by the concerns and opinions, and 
    the response by either the tailoring score or preferred response, depending on the setting. 
For the evaluation prompt, we use zero-shot prompting in both settings. The temperature is set to 1 for the fine-tuned LLaMA model. 
To extract the model's answer, we only use the first sentence of the response. 
For the absolute scoring setting, we search for numeric values in the sentences. 
For the relative preference setting, we look for either \textit{A} or \textit{B} in the sentence.

\subsection{Performance}
\Cref{tab:auto_eval_results} summarizes each automatic evaluator's performance on \ourdataset.
The fine-tuned BERT model outperforms all other models.
meaning the student model generalized better in this case.
This result mirrors similar observations made about self-distillation, which has been shown to have a regularizing effect \cite{furlanello2018born,mobahi2020self}.
Absolute scoring performed better in both GPT-4-Turbo and BERT, while relative preferences were more accurate in Llama-2. 
We also found that continuing finetuning with the human-labeled data after first training on LLM labeled data improved performance for all our models.
In the relative preference setting, this alone improved Llama-2 performance from $69.1\%$ to $73.9\%$.
We observed that Llama-2 finetuning is sensitive to hyperparameters, but due to limited time and resources we were not able to fully tune these to improve beyond the performance of the BERT models.

Incorrect predictions of \textit{Equal} should not be penalized the same as mismatches between \textit{Response A} and \textit{Response B}, so we compute 
a ``lenient'' version of accuracy:
we ignore these errors by removing all \textit{Equal} predictions before computing accuracy (treating them as \textit{Abstain}s).

\begin{table}[h]
        \centering
        \small
        \begin{tabular}{lllll}
        \toprule
        Model                          & Setting        & Dev     & Test \\
        \midrule
        \multirow{2}{*}{GPT-4-Turbo}   & Pref.          & 58.5    & 69.3  \\
                                       & Score          & 65.5    & 76.5  \\
        \multirow{2}{*}{Llama-2 (13B)} & Pref.          & \textbf{69.8}    & 73.9  \\
                                       & Score          & 62.7    & 68.7  \\
        \multirow{2}{*}{BERT}          & Pref.          & 62.0    & 77.0  \\
                                       & Score          & 65.0    & \textbf{80.8}  \\
        \bottomrule
    \end{tabular}
    \caption{\textbf{Accuracy (\%) of Automatic Evaluators on Dev and Test Sets}.
    Evaluators are built in two settings: predicting relative preferences (Pref.) between two input responses, or predicting absolute scores (Score) for a single input response.
    Accuracy is then computed on Dev and Test sets as the percentage of pairwise preferences the model correctly ranks.
    }
    \label{tab:auto_eval_results}
\end{table}

Training on relative preferences sometimes performs worse than training with absolute scores, despite the higher quality data in the human-labeled partition (\S\ref{sec:agreement}).
This could potentially be attributed to a loss of information: for both the case where \textit{rA} was \emph{much} better than \textit{rB} and the case where it was only slightly better, the eventual label in the relative preference setting is the same, while absolute scoring distinguishes these.
Therefore, there is a potential tradeoff between annotation quality (best annotation type for crowdsource workers) and training efficiency (best annotation type for models) that could be studied further.
On the other hand, forcing a decision when two responses are indistinguishable may be a new source of noise.

\section{Results}

To better explore the usefulness of \ourdataset towards improving our understanding of tailored responses, we investigate what models, strategies, and opinions work best.
This section describes a series of analyses looking at factors to consider when generating responses tailored to CGOs, which may be of interest to NLP researchers as well as public health professionals working on vaccine hesitancy.

\begin{figure}[h]
   \centering
   \includegraphics[width=.48\textwidth]{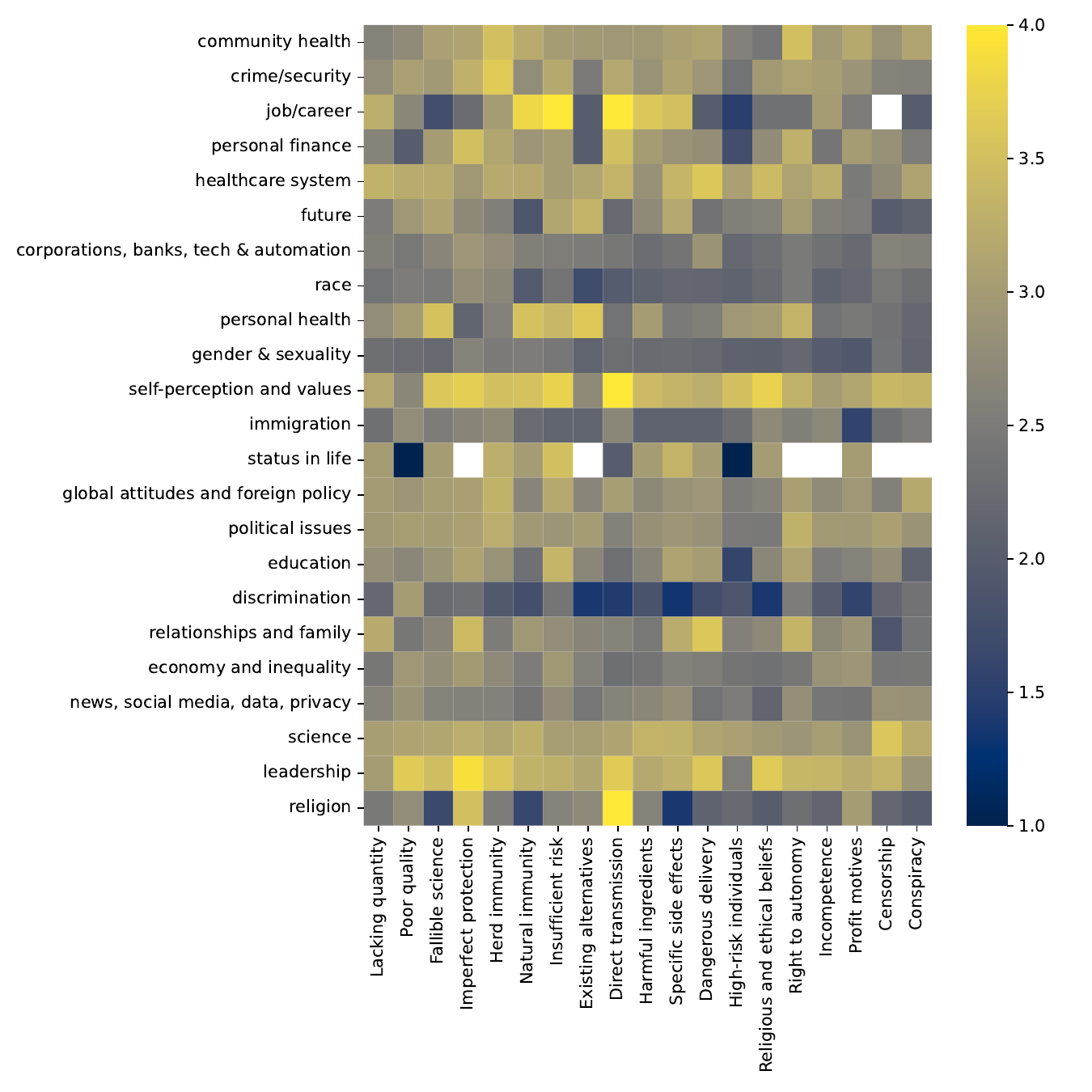}
   \setlength{\abovecaptionskip}{-10pt}
   \caption{
        \textbf{Heatmap of mean scores by LLM evaluation for responses answering a concern (horizontal axis) while tailoring to a CGO (vertical axis)}.
        Brighter colors indicate higher scores, while white squares are nulls that were not sampled during annotation.
        Religion, while an opinion topic that scores poorly in our testing, seems to provide useful opinions for tailoring when focusing on the \textit{Direct transmission} concern (see \Cref{apx:religion_for_direct_transmission} for an example output).
    }
    \label{fig:heatmap}
\end{figure}

\subsection{Opinion Selection}
\begin{figure}[h]
   \centering
   \includegraphics[width=.48\textwidth]{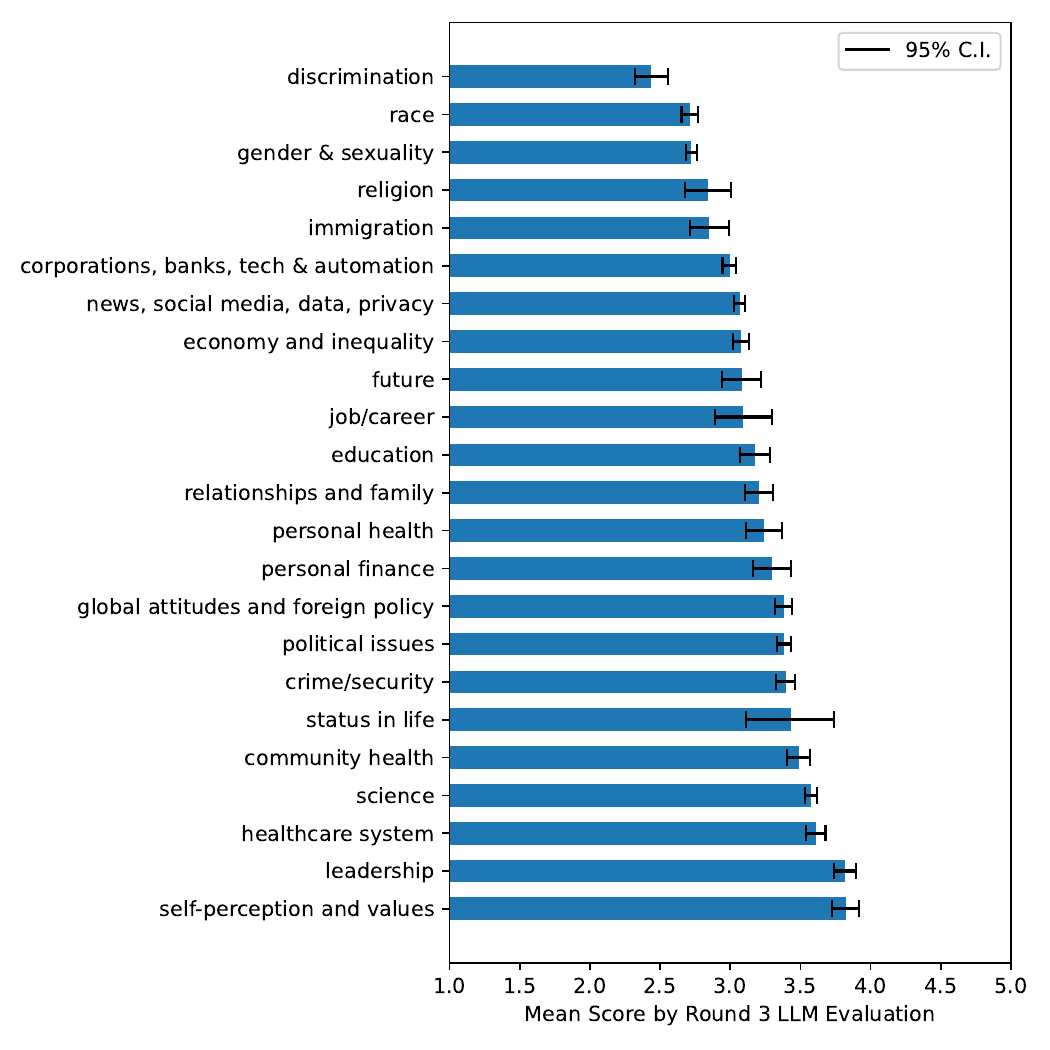}
   \setlength{\abovecaptionskip}{-10pt}
   \caption{
        \textbf{Comparison of mean response quality for each CGO, aggregated by topic}.
        Notice that potentially controversial and problematic topics such as discrimination, race, or religion are bad targets for tailoring.
        The implications of this result is that using divisive topics to establish common-ground may be less useful, and using less polarized topics (self-perception) for example can result in stronger overall scores.
    }
    \label{fig:best_opinions}
\end{figure}

A driving motivation for this work is to allow analysis as to which opinions are fruitful for tailoring vaccine messaging on.
To this end, we conduct an analysis of which opinions led to the best tailored responses by examining the mean scores (\Cref{fig:best_opinions}) within topic clusters of opinions.
We use the 24 topics proposed and annotated by \citet{santurkar_2023_opinions} for the Pew Research's American Trends Panels questions.

For the analysis in \Cref{fig:best_opinions}, we collect all candidate tailored responses in the round-3 data if they tailor to an opinion associated with our topic of interest.
The GPT-4-Turbo assigned score for each such responses is then computed by drawing 100 predictions and averaging;
the final reported score is a second average taken over each candidate response's score.
To determine a 95\% (bootstrap) confidence interval, we repeat this process 10,000 times for each topic by drawing from the candidate responses with replacement.
The analysis is repeated for each of the 24 topics.

However, some opinions may be better suited for use in a small subset of concerns.
We therefore investigate the response quality when the CGO belongs to each topic by repeating the process on    each concern category in the VaxConcerns taxonomy \citep{stureborg2023vaxconcerns}.
The mapping between concerns and these concern categories is discussed further in \S\ref{sec:concerns_and_opinions}.

\Cref{fig:heatmap} shows a visualized heatmap of the mean scores of each <opinion-topic, concerns-category> combination.
These results indicate that topics can indeed be better suited for tailoring responses to some concerns than others.
For example, job/career opinions do very well on average with the \textit{Insufficient risk} and \textit{Direct transmission} concerns (approx. 4/5), but quite poorly with concerns regarding \textit{Existing alternatives} (approx. 2/5).
To improve automatic vaccine messaging, dynamically selecting the right opinions when addressing a concern could be a key strategy to improving response quality from models.
We note that broad concerns (Level 1 in the VaxConcerns taxonomy) are generally easier to address than specific concerns (Level 2).
This may be because specific concerns are harder to creatively link with a given CGO, while broader concerns offer more potential ways to relate the two topics.

\begin{figure}[h]
   \centering
   \includegraphics[width=.48\textwidth]{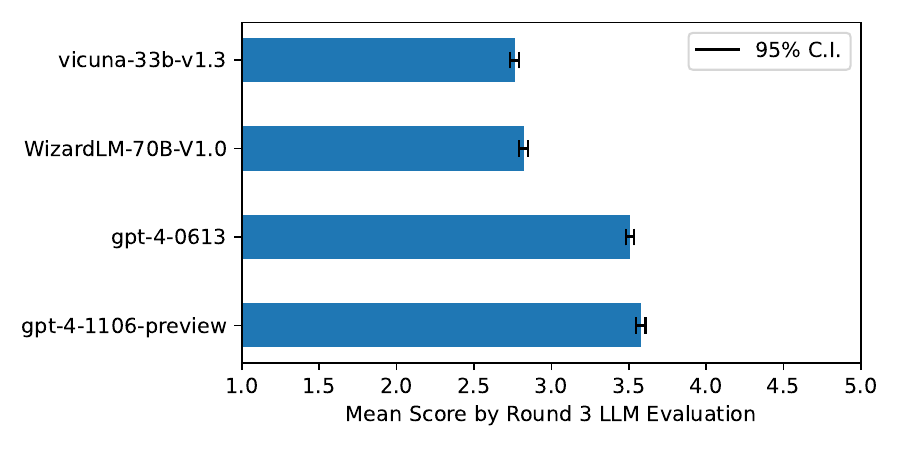}
   \setlength{\abovecaptionskip}{-10pt}
   \caption{
        \textbf{Comparison of Mean Response Quality by each Model in the LLM-annotated train set partition}.
        All differences in the figure are statistically significant.
        Confidence intervals are computed through bootstrap sampling.
        Each model is evaluated across approximately 4,000 generated responses each to randomly sampled concern and opinion statements.
        We see GPT-4-Turbo produces the best tailored responses on average, just ahead of GPT-4.
        Open-source models still lag far behind, despite using the largest possible model sizes on our hardware.
    }
    \label{fig:best_models_round3}
\end{figure}

\subsection{Model}
\label{sec:model_results}
\begin{figure}[h]
   \centering
   \includegraphics[width=.48\textwidth]{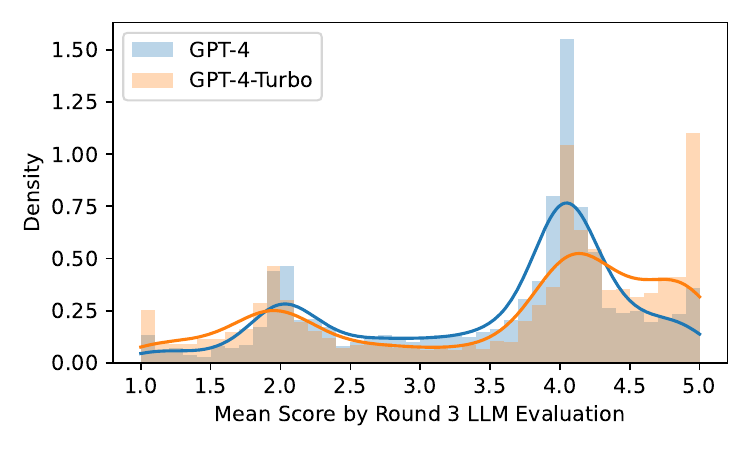}
   \setlength{\abovecaptionskip}{-10pt}
   \caption{
        \textbf{Score distributions for GPT-4 and GPT-4 Turbo in round-3 dataset partition}.
        GPT-4-Turbo has a slightly better mean score than GPT-4 (3.58 vs 3.51).
        The distributions of response quality is generally comparable between the models, with GPT-4-Turbo producing notably higher proportion of scores around 4.
        Note that the distribution of scores seems to be bimodal, potentially due to scores either successfully finding or not finding a useful linking idea between the concern and CGO.
    }
    \label{fig:score_distr_round3}
\end{figure}

We benchmark the performance of several major LLMs through the large-scale data collection in \ourdataset.
\Cref{fig:best_models_round1} shows a breakdown of model performances as determined by human annotators.
While these results are helpful, further analysis requires larger datasets, for which we use the LLM based annotations of the round 3 partition (\Cref{fig:dataset_partitions}) as shown in \Cref{fig:best_models_round3}.
GPT-4-Turbo is shown to perform the best on \ourdataset, and is subsequently used for an analysis as to best prompting strategies in \S\ref{sec:prompting_results}.
Additionally, \Cref{fig:score_distr_round3} shows a closer look of exactly where GPT-4-Turbo performs better than the next best model, GPT-4 by examining the distribution of scores assigned to tailored responses written by each model.

\subsection{Prompting}
\label{sec:prompting_results}
We determine the best prompting method using the best model outlined in \S\ref{sec:model_results}.
We compare configurations of the 3 prompt dimensions described in \S\ref{sec:prompting_method}.
Roles are compared against each other in \Cref{fig:best_roles_round3}, and the score distribution for two selected roles is plotted in \Cref{fig:politician_vs_health_expert}, both in \Cref{apx:roles_comparisons}.
\Cref{fig:score_distr_cot_vs_non_cot} in \Cref{sec:cot_or_not} describes the difference between Chain-of-Thought (CoT) prompting and standard prompting (non-CoT). 
Standard prompting is significantly better than CoT.
For GPT-4-Turbo, the best prompting strategy is to use the \textit{Health Expert} role with guidelines and non-CoT prompting.

There is not a statistically significant improvement in model responses through use of the guidelines,
although our qualitative observations had indicated that guidelines led to responses that better adhered to the principles included by the guidelines.
However, we investigate the case where guidelines are sampled or are left out entirely, and there could be more work to determine if a subset of the guidelines significantly improves the response quality.

\subsection{Expert Evaluation}
We asked 3 senior public health experts to rate and comment on the quality of the generated responses to validate their potential usefulness to a vaccine navigator.
They collectively rated 60 unique pairs of tailored responses, and were asked to select their preference and give comments describing their general impressions for who these responses are appropriate to.
All responses were generated from the strongest model in \Cref{sec:model_results}, but in each pair one was a randomly chosen response, while the other was filtered as the best of 20 random generations by the BERT-based automatic evaluator from \Cref{sec:auto_eval}.

We were not able to find a statistically significant difference between the filtered response and the randomly chosen ones in this sample.
This may be due to the model's overall strength, as evidenced by the claimed usefulness of the responses according to the experts.
Experts overall described the responses as ``very high quality'' and noted these responses could ``easily be used by vaccine navigators''.
Responses were sorted into four categories for overall quality: Low, Medium-Low, Medium-High, and High.
Only $5\%$ of responses were categorized as ``Low'' quality, $20\%$ were ``Medium-Low'' $20\%$ were ``Medium-High'', and $55\%$ were ``High''.

Further, for $87.5\%$ of responses, they said they would be useful to a vaccine navigator.
For $55\%$ of responses, they said they would even be comfortable with a patient reading the response.
The main issues identified in responses were to do with being too technical, most often meaning they did not recommend showing these to patients.
This seems to indicate that the responses by the strongest identified model in this work may be good enough to aid a vaccine navigator in their work, demonstrating the potential direct application of the framework.

\section{Conclusion}
We introduce \ourdataset, a comprehensive and high-quality dataset for training and evaluation of tailoring vaccine interventions to common-ground opinions.
We benchmark several major LLMs, finding that GPT-4-Turbo best tailors responses to CGOs. 
We build evaluation metrics on top of this dataset to allow cheap and accurate evaluation of models.
Finally, we analyze which opinions are better suited to tailor vaccine interventions with, and provide recommendations for which opinions to select for specific concern categories.

\section*{Limitations}

\textit{Finding opinions for tailoring}.
In this work we present methods to tailor on common-ground opinions.
We assume that we are given these opinions and the intended audience believes or strongly believes in them.
However, identifying beliefs is a difficult task on its own and requires further research.
The easiest way to confirm the audience holds the opinion is to survey them, but doing so may affect further communication in other ways.

\textit{Crowdsource Workers' Biases}.
We attempt to define annotation tasks that should be ``objective'' regardless of who is labeling (thereby our focus on inter-annotator agreement).
However, each annotator brings in their own personal biases.
Opinions that seem questionable or off-putting to one annotator may influence their ranking on that example.

\textit{Tailoring versus Engagement and Subjectivity}.
One motivation for this work is to provide a framework for generating candidate responses in vaccine interventions.
To create a well-defined task, we measure how well concepts are related (linked) in writing.
What we do not directly measure is how engaging, persuasive, or applicable each response is to the actual person reading it.
Such tasks are much more subjective, and require extensive invitation of participants from diverse backgrounds in order to ensure solid research findings.
Individual identities are highly inter-sectional, and it can be hard to recruit participants for whom vaccine messaging is intended for.

\textit{Diversity of Generated Responses}.
We attempt to create a large diversity in responses by sampling many different models, under varied instructions, and with a variety of concerns and opinions.
However, we are still restricted in the diversity we are able to generate on many dimensions.
We begin to explore this further in \ref{apx:response_lens}.
Future work could explore tasks such as tailoring longer documents, or technical writing.

\textit{Use of Automatic Evaluation}.
It has been pointed out that LLM evaluators have remaining challenges such as poor performance on higher-quality models \citep{shen-etal-2023-large} or a preference for text generated by itself, likely due to a bias in low-perplexity examples \citep{stureborg2023autoeval}.
Our work therefore uses other methods of evaluation as well, such as direct accuracy comparisons with human labeled data, and averages over several evaluations as suggested in \citet{stureborg2023confidence}.

\section*{Ethics statement}
By the nature of opinions and vaccine misinformation, the research in this paper comes with important ethical considerations.
Some opinions from the PEW Research surveys are bound to be controversial and
asking generative models to assume such opinions as truth may produce potentially harmful responses.
Thus, to use tailored responses, there is a great deal of responsibility in selecting common-ground opinions to tailor on to avoid perpetuating harm.
Some of the CDC-sourced principles for vaccine messaging address these issues, such as the principle that ``CDC's communication is empathetic, respectful, non-judgmental, and never arrogant toward others''.
The systems which our work begins to explore are not yet performing at a high level of tailoring.
Many responses are completely artificial and don't seem very persuasive.
However, our results indicate that producing well-tailored responses is possible.
Given the potential uses (and misuses) of such systems which this work begins to explore, it is paramount that more research is done on safety and harm within this field.

\section*{Acknowledgements}
We are very grateful to Walter Orenstein, Robert A. Bednarczyk, and Lavanya Vasudevan for their help with the evaluation of response quality.
A special thanks to Christopher Kjellqvist for many useful discussions on signals for personalization and persuasion, and the whole NLP group at Duke for their helpful feedback and input on the project.
This work was supported by NSF award IIS-2211526.
Thank you to the anonymous reviewers for their feedback.

\bibliography{anthology,custom}
\bibliographystyle{acl_natbib}

\appendix
\onecolumn

\section{Example Responses}
\label{apx:example_responses}

\begin{figure}[h]
   \centering
   \includegraphics[width=0.9\textwidth]{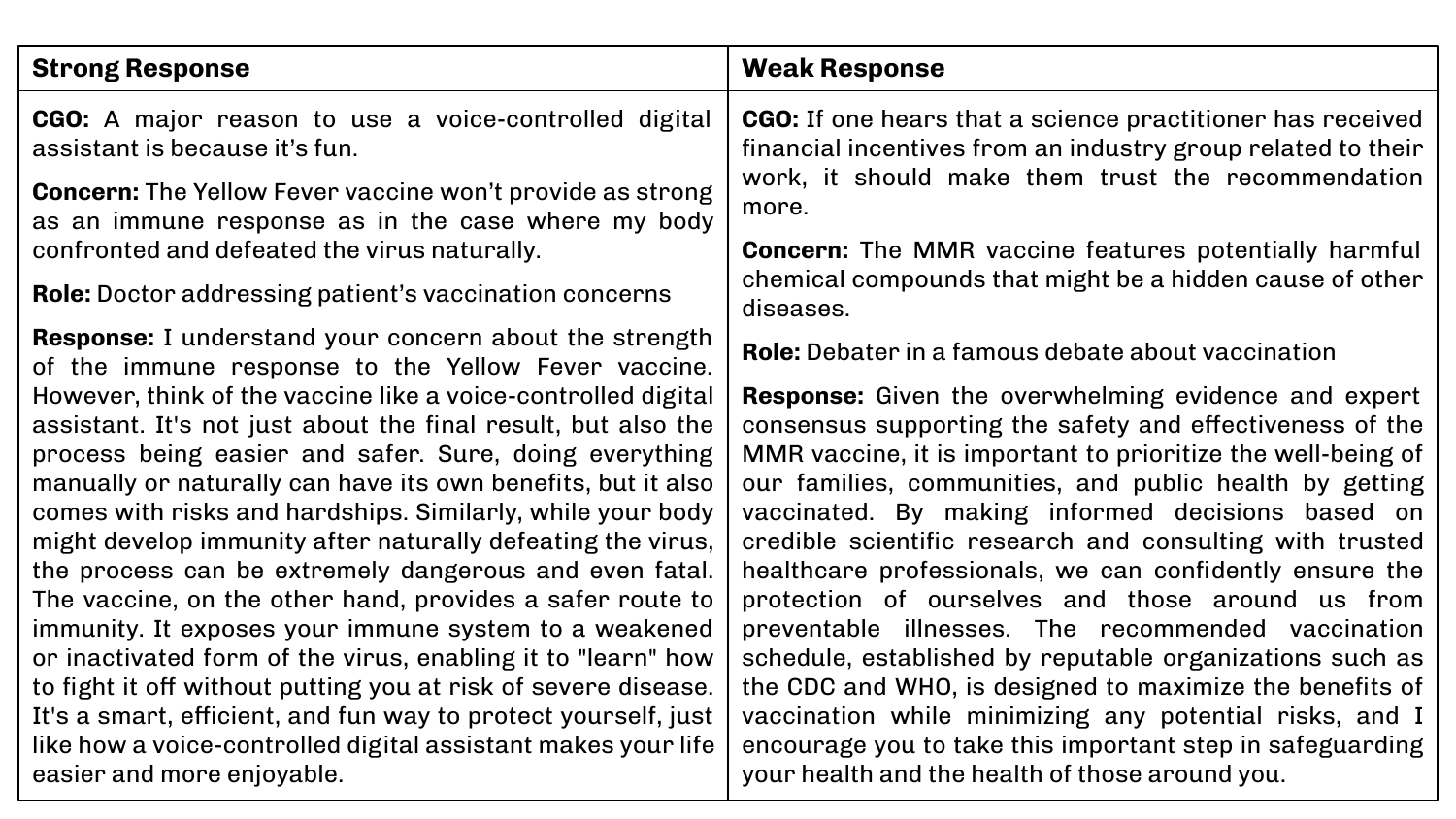}
   \setlength{\abovecaptionskip}{2pt}
   \caption{
        \textbf{Example of Strong vs Weak Response} The strong response leverages a powerful analogy using the CGO to address the concern, making the crucial point that despite the common end result of vaccination and natural immunity, vaccination makes the process easier and much safer. The weak response is generic, failing to both address the specific concern regarding potentially harmful chemicals as well as using the CGO in its argument in any meaningful way.
    }
    \label{fig:example_responses}
\end{figure}

\section{Response Lengths in \ourdataset}
\label{apx:response_lens}

\begin{figure*}[h]
   \centering
   \begin{subfigure}{0.33\textwidth}
    \centering
    \includegraphics[width=\textwidth]{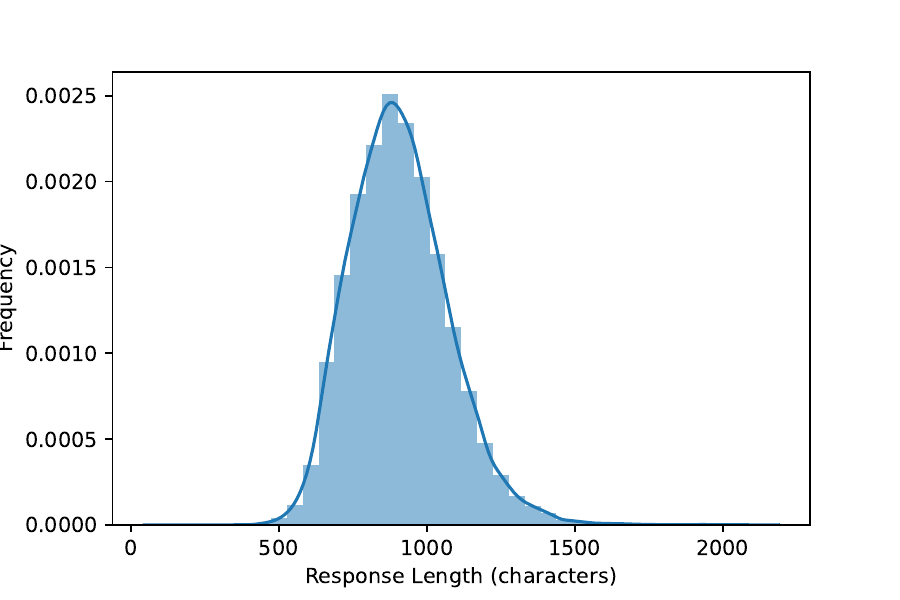}
  \end{subfigure}%
  \begin{subfigure}{0.33\textwidth}
    \centering
    \includegraphics[width=\textwidth]{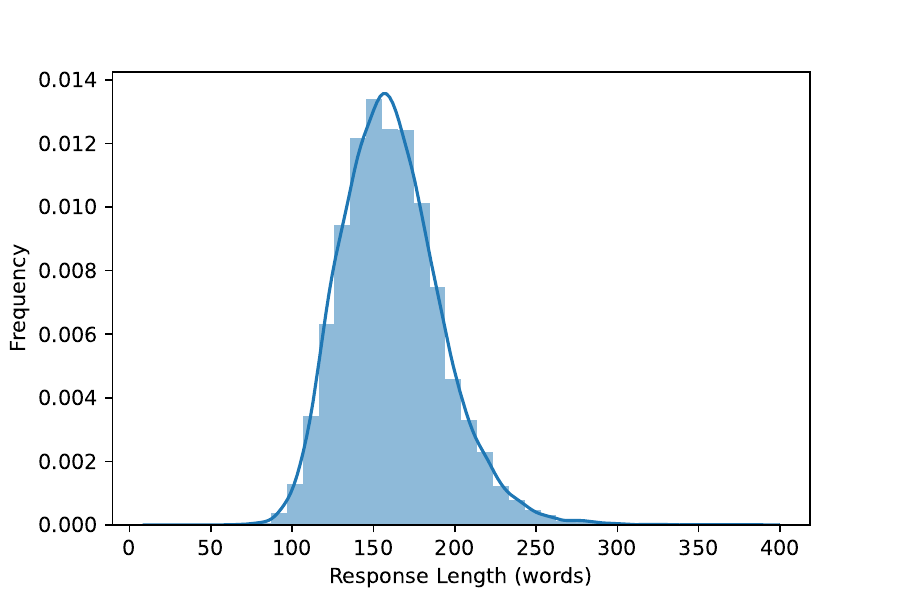}
  \end{subfigure}
  \begin{subfigure}{0.33\textwidth}
    \centering
    \includegraphics[width=\textwidth]{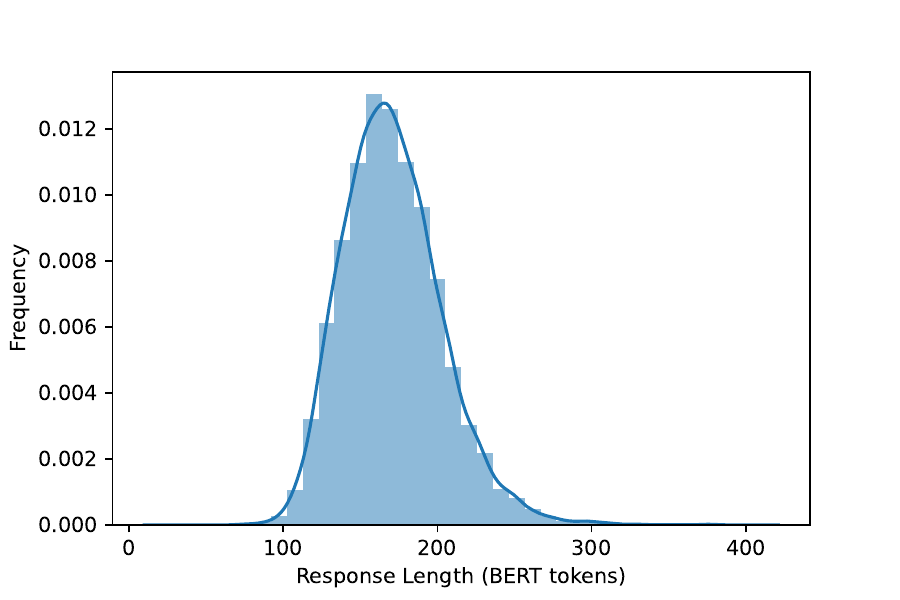}
  \end{subfigure}
   \caption{
        \textbf{Distribution of length of responses in \ourdataset}.
        Our responses cover a range of short to medium length passages.
        All generative models were instructed to answer the concern in ``a short paragraph''.
        Responses are generally within the token limits of models such as BERT, but not strictly bound to such ranges.
        The shortest response is 27 tokens, and the longest response is 421 tokens.
        Future work may investigate this task while allowing for longer response lengths.
    }
\end{figure*}

\section{Annotation Platform}
\label{apx:annot_platform}
We use a custom-built annotation platform to prompt responses from annotators.

\begin{figure}[h]
   \centering
   \includegraphics[width=1.0\textwidth]{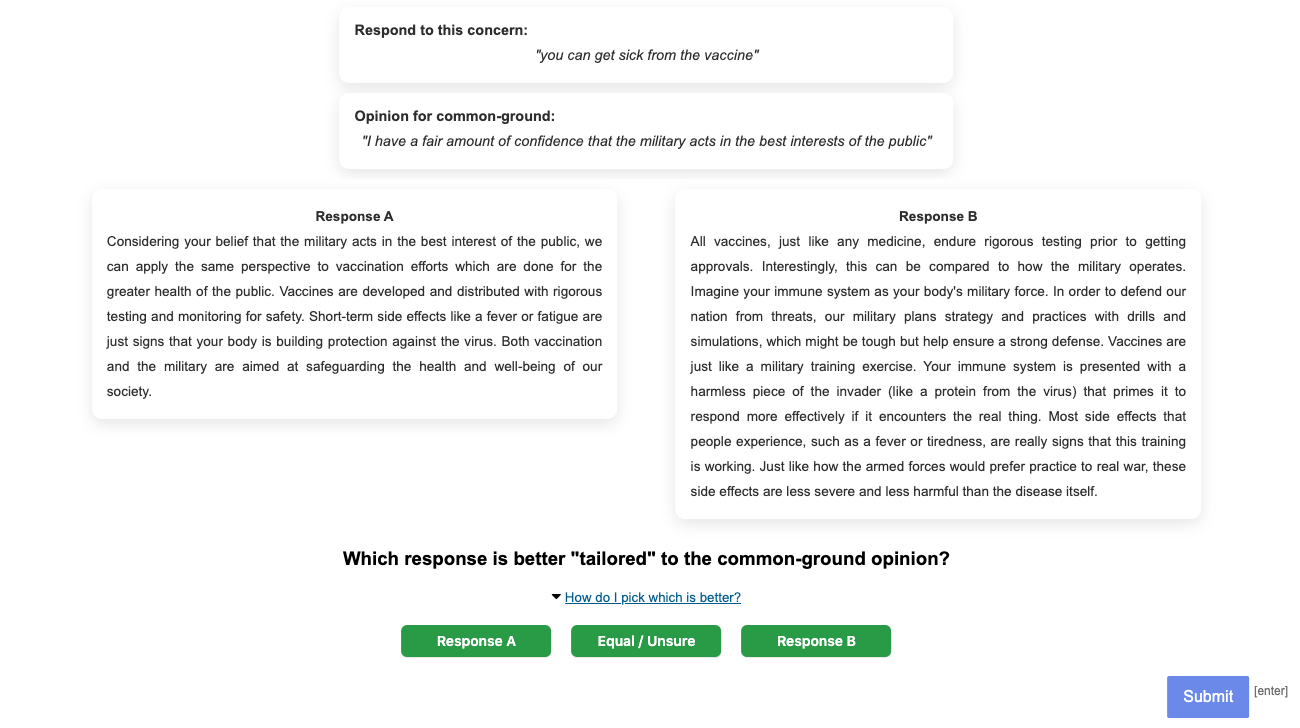}
   \setlength{\abovecaptionskip}{-10pt}
   \caption{
        \textbf{Annotation Task Example}.
        The annotation interface displays the specific vaccine concern, the common-ground opinion, and two adjacent responses to compare and choose between.
    }
    \label{fig:annotation_task}
\end{figure}

\begin{figure}[h]
   \centering
   \includegraphics[width=0.6\textwidth]{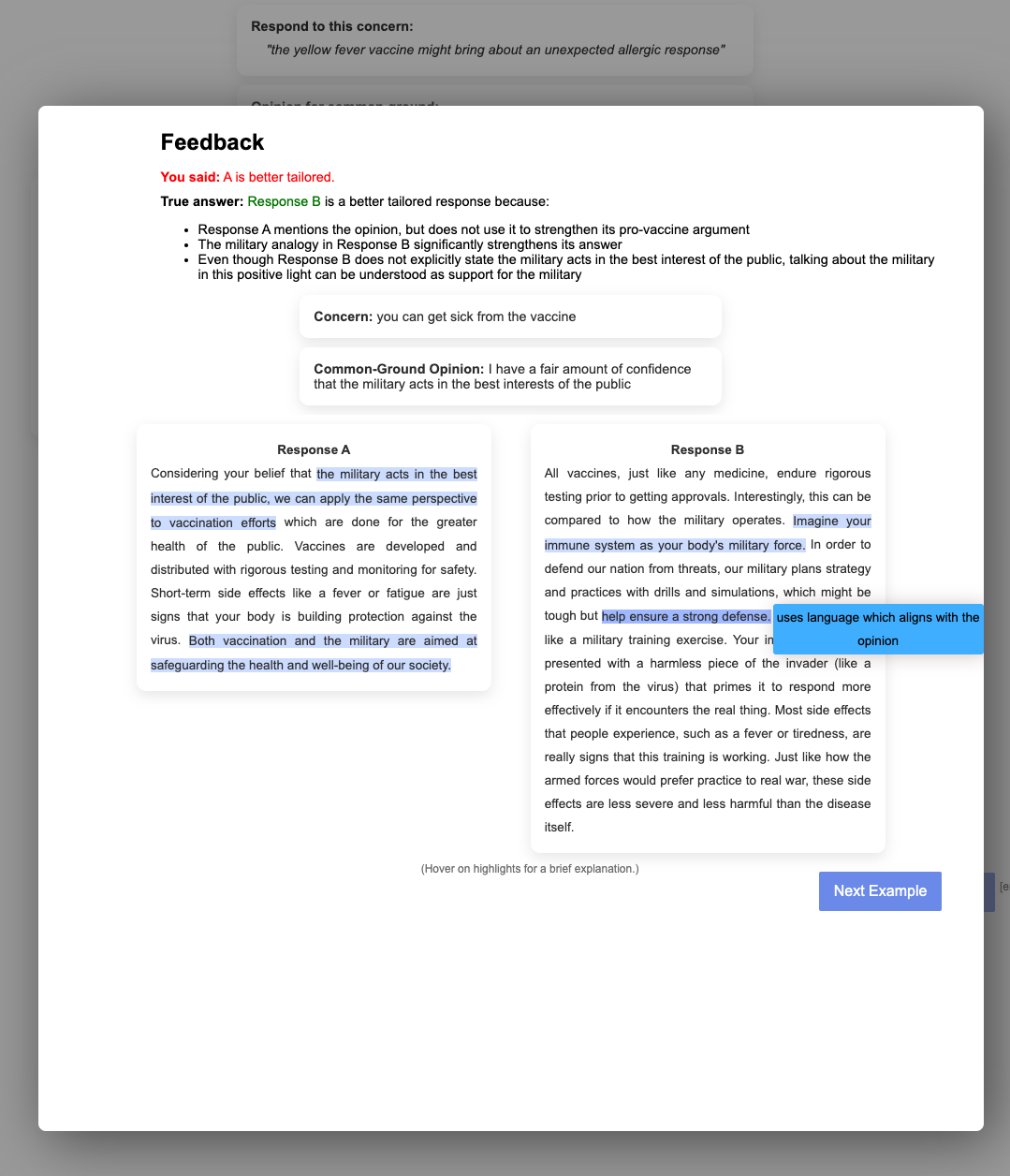}
   \setlength{\abovecaptionskip}{7pt}
   \caption{
        \textbf{Annotation Tutorial Example}.
        The interface displays an instance of the interactive tutorial through which annotators learn how to choose between the two responses. They attempt trial questions and are given feedback. 
    }
    \label{fig:annotation_tutorial}
\end{figure}

\section{Score Distributions Between GPT-4-Turbo and Open-source Models}

\begin{figure}[h]
   \centering
   \begin{subfigure}{0.5\textwidth}
    \centering
    \includegraphics[width=\textwidth]{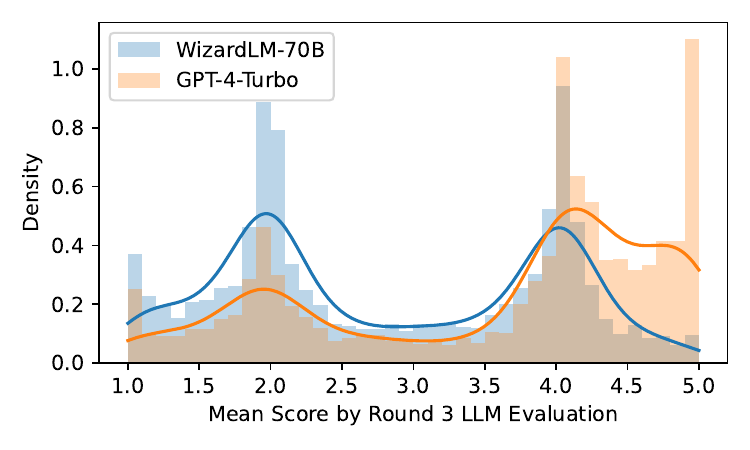}
  \end{subfigure}%
  \begin{subfigure}{0.5\textwidth}
    \centering
    \includegraphics[width=\textwidth]{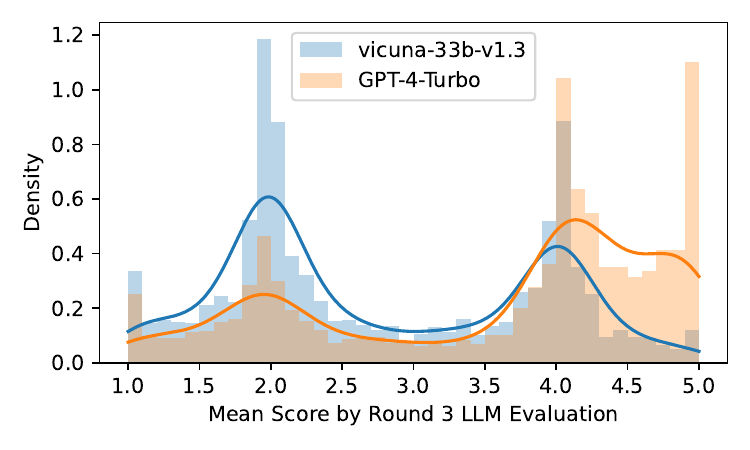}
  \end{subfigure}
   \setlength{\abovecaptionskip}{-10pt}
   \caption{
        \textbf{Score distributions for GPT-4 two Open-Source Models}.
        Well-performing models have distributions shifted towards higher scores (right).
        GPT-4-Turbo produces a higher frequency of responses scored between 4-5 than the two other models, while both models produce more responses scored around 2 than GPT-4-Turbo.
    }
\end{figure}

\section{Models}
\label{apx:models_used}

We use several family of models and several different checkpoints and sizes in various experiments.
\Cref{tab:models_used} lists all such model checkpoints and notes in which experiments they were used.

\begin{figure}[h]
   \centering
   \includegraphics[width=.48\textwidth]{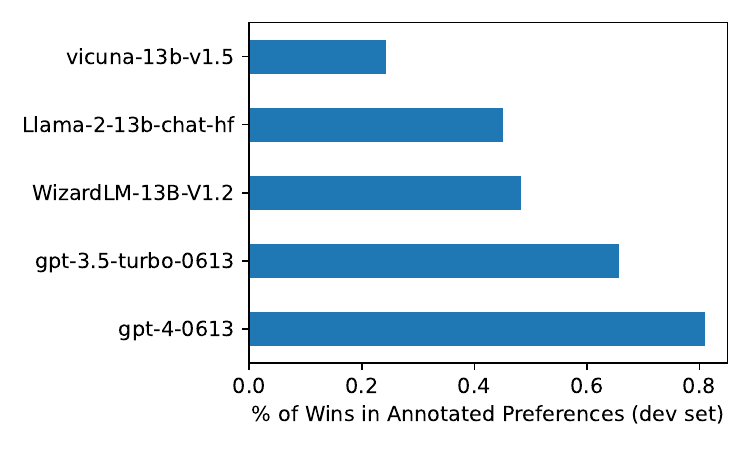}
   \caption{
        \textbf{Comparison of model win rates when paired up against one another during round-1 annotation}.
        This dataset offers us limited opportunity at measuring differences due to the cost of annotating enough data.
        However, since the number of models explored in the dev set was just 5, we can determine GPT-4 produced the highest quality responses as determined by annotators.
    }
    \label{fig:best_models_round1}
\end{figure}

\section{Qualitative Analysis of Generated Response Quality}
\label{apx:qualitative_analysis}

\subsection{Statistics}
We looked at 700 sample responses out of a total of 1546 in our unguided responses dataset. Within this sample, we found 21 instances of model failure, demonstrating a 3\% failure rate.\\ 

\subsection{Failure Modes}
Within our sample, there were several distinct failure modes that we identified with varying frequencies:
\begin{enumerate}
    \item The model directly or indirectly assumes the identity of a human 13 times.
    \item The model assumes an identity for the person it is responding to 2 times.\\
    For instance, for a general CGO regarding athletes, the model assumes the person it is responding to is also an athlete. 
    \item The model explicitly disagrees with the CGO 1 time.
    \item The model responds to a different vaccine concern than that which is provided in the prompt 2 times.
    \item The model makes a factually incorrect or ambiguous statement 1 time. 
    \item The model crafts a response that contains a template element like “hey [friend’s name]” 2 times.\\
\end{enumerate}

\subsection{Notable Cases}
\label{apx:religion_for_direct_transmission}
\begin{enumerate}
    \item \textit{Role: Twitter User}\\
    We observed that when the model was prompted to respond from the role of a Twitter user, it often demonstrated strong personal agreement, a quality that is not representative of other responses. For instance, the model would start its response with the statement “I completely agree with you.” We observed this case 6 times out of 54 cases (11.1\%) of the Twitter user role within our sample.
    \item \textit{CGO: Religion}\\
    We observe that while religion is an opinion topic that generally scores poorly in our testing, the model is able to create strong responses when tailoring it to the concern regarding direct transmission of the virus from the vaccine.\\
    \textit{Example Response}\\
    \textbf{CGO:} Being an evangelical Christian helps a lot in people's ability to get ahead in our country these days.\\
    \textbf{Concern:} Getting the COVID-19 vaccine might directly infect me with the virus it's supposed to protect me from.\\
    \textbf{Response:} As an evangelical Christian, you may find comfort in the fact that many faith leaders have encouraged their communities to get vaccinated, seeing it as an act of love for one's neighbors and a way to protect vulnerable populations. It's important to understand that COVID-19 vaccines do not contain live virus and cannot infect you with the disease. Instead, they use a small piece of the virus's genetic material or a weakened or inactivated form of the virus to stimulate your immune system to produce antibodies. These antibodies help your body recognize and fight the actual virus if you're exposed to it in the future. By getting vaccinated, you not only protect yourself but also contribute to herd immunity, which helps safeguard those who cannot receive the vaccine due to health reasons. In this way, getting vaccinated aligns with your shared perspective and can be seen as a way to live out your faith by serving and protecting others.\\
\end{enumerate}

\subsection{Characteristics of our responses}
In general, we can divide our common-ground opinions into three categories with respect to their proximity of subject matter to vaccination: 
\begin{itemize}
    \item Directly related, including topics such as healthcare or public health.
    \item Somewhat related, including topics such as food, well-being, or economic growth.
    \item Unrelated, including topics such as guns, personal relationships, or race relations.
\end{itemize}
For our specific task of adapting a vaccine response to a given common ground opinion, we find that the model is capable of generating strong responses for topics spanning all three categories.\\
Conversely, weaker responses usually take 3 forms: 
\begin{itemize}
    \item The model attempts to incorporate the CGO, but fails to do so effectively.
    \item The model acknowledges the CGO, but does not use it in its response. 
    \item The model ignores the CGO.
\end{itemize}
Examples of responses we characterize as \textit{strong} and \textit{weak} for this task are given below.\\ 

\subsection{Examples of strong responses}
\begin{enumerate}
    \renewcommand{\labelenumi}{\Roman{enumi}.}
    \item \textit{Directly related}\\
    \textbf{CGO:} Medical research scientists care about the best interests of the public all or most of the time.\\
    \textbf{Concern}: My elderly father, with his weak immunity, may experience severe health risks from vaccines.\\
    \textbf{Role:} Twitter User\\ 
    \textbf{Response}: It's heartening to know you trust the integrity of medical research scientists, as their work is indeed guided by a commitment to public health and safety. When it comes to your elderly father with weaker immunity, it's natural to be concerned about potential risks. However, vaccines are specifically designed with vulnerable populations in mind and, in fact, are crucial for them. As someone with a delicate health situation, your father may benefit significantly from the added protection vaccines offer against serious diseases. Most vaccines undergo rigorous testing to ensure they're safe for people like him, and any risks are typically far outweighed by the risk of the disease itself. It's worth discussing with his healthcare provider about the best vaccines for his situation—this personalized approach ensures his care is aligned with his needs and gives him the best chance to stay healthy and strong. \#VaccinesWork \#PublicHealth
    
    \item \textit{Somewhat related}\\
    \textbf{CGO:} The use of data or information by a social media company to monitor its users' posts for signs of depression, in order to identify people who are at risk of self-harm and connect them to counsel, is acceptable.\\
    \textbf{Concern:} The yellow fever vaccine being enforced as mandatory, without respecting my right to decide what is best for my body, is an overstep by the authorities.\\
    \textbf{Role:} Parent talking to another parent\\
    \textbf{Response:} I completely understand your concerns about personal autonomy and the right to make decisions about your own body. However, consider how you're comfortable with social media companies using data to prevent self-harm. Vaccination works on a similar principle - it's a preventive measure aimed at protecting not just you, but also those around you. Just as social media monitoring can help prevent a person from self-harming, vaccines prevent the spread of serious diseases like yellow fever. By choosing to vaccinate, you're not only protecting yourself, but also contributing to the larger community's health through herd immunity. It's not about infringing on personal rights, but about safeguarding public health. I encourage you to discuss your concerns with a healthcare professional who can provide personalized advice. Remember, the goal of vaccination is to protect, not to harm or control.
    
    \item \textit{Unrelated}\\
    \textbf{CGO:} Being a supporter of the National Rifle Association (NRA) is important.\\ 
    \textbf{Concern:} Can the Yellow Fever vaccine offer full protection, given reports of its imperfect efficacy rate?\\
    \textbf{Role:} Diplomat representing country at press conference asked to provide vaccine interventions\\
    \textbf{Response:} I understand your attachment to the principles of the NRA, which include the preservation of individual freedom and the right to protect oneself. I'd like to draw a parallel between that notion and our efforts in vaccination. The Yellow Fever vaccine, like any other vaccine, is a tool for self-protection. While it's true that it doesn't guarantee 100\% protection, it significantly reduces the risk of contracting the disease. It's similar to how owning a firearm doesn't guarantee complete safety, but it does provide an essential line of defense. In both cases, we are taking proactive measures to protect ourselves, our families, and our communities. We encourage everyone to get vaccinated not just for individual protection, but to strengthen our collective defense against the virus, just as the NRA encourages responsible firearm ownership for the safety of all.\\
\end{enumerate}

\subsection{Examples of weaker responses}
\begin{enumerate}
    \renewcommand{\labelenumi}{\Roman{enumi}.}
    \item \textit{Fails to incorporate CGO effectively}\\
    \textbf{CGO:} Today, a wide range of information about people is searchable online, and it is more important for people to have the ability to discover potentially useful information about others.\\
    \textbf{Concern:} are yellow fever vaccines really necessary or just another money-making scheme?\\
    \textbf{Role:} AI assistant tasked with addressing user inquiries\\ 
    \textbf{Response:} Given your perspective on the importance of accessible information, it's crucial to understand the role of yellow fever vaccines in global health. Yellow fever is a serious viral disease that's prevalent in certain parts of Africa and South America, and the vaccine is not a mere money-making scheme. It's a proven and effective measure to prevent the disease, which can be fatal. The World Health Organization recommends the vaccine for people traveling to these regions. The cost of the vaccine is not only about making a profit, but also covers research, development, and distribution. In the spirit of transparency and information accessibility, it's important to note that the benefits of this vaccine far outweigh the financial implications.

    \item \textit{Acknowledges CGO, but does not use it in its argument}\\
    \textbf{CGO:} Women claiming they have experienced sexual harassment or assault in the workplace when it hasn't actually happened is a major problem.\\
    \textbf{Concern:} The body's own immune system, if properly nurtured, can fend off any infections better than vaccines can.\\ 
    \textbf{Role:} Active Reddit User\\
    \textbf{Response:} While it's essential to recognize the importance of addressing concerns about sexual harassment and assault, it's also crucial to prioritize our health and well-being. Vaccines have been proven to be safe and effective in protecting individuals against various diseases, and they also contribute to the greater good by promoting herd immunity. By getting vaccinated, you're not only safeguarding yourself but also helping to prevent the spread of diseases within our community. Your health and the well-being of those around you are both important, and taking the step to get vaccinated is a proactive way to ensure both are protected.\\
    
    \item \textit{Ignores CGO}\\
    \textbf{CGO:} in general, it is essential for someone in high political office to work well under pressure.\\
    \textbf{Concern:} Natural remedies, like boosting your immune system to fight viruses, are a viable alternative to the HPV vaccine.\\
    \textbf{Role:} Politician delivering a speech on vaccine-related subjects\\ 
    \textbf{Response:} I understand that you may have concerns about the HPV vaccine, but I want to assure you that it is a safe and effective way to protect yourself and your loved ones from serious diseases. The vaccine has been extensively tested and has been proven to prevent cervical, anal, and other types of cancer, as well as genital warts. By getting vaccinated, you not only protect yourself, but you also help to prevent the spread of these diseases in your community, which is especially important for vulnerable individuals such as young children and the elderly. Additionally, getting vaccinated is an important way to take control of your health and well-being, and it's a simple and effective way to protect yourself from diseases that can have serious consequences. I hope this information has been helpful in addressing your concerns, and I encourage you to make an informed decision to get vaccinated.\\
\end{enumerate}

\subsection{Example of failure}
\begin{enumerate}
    \renewcommand{\labelenumi}{\Roman{enumi}.}
    \item \textit{Model Hallucinates}\\
    \textbf{CGO:} What happens to Hispanics in the country overall affects my own life a lot.
    \textbf{Concern:} If we rely too much on the HPV vaccine, we may not thoroughly explore other effective preventive measures.\\ 
    \textbf{Role:} Someone speaking to their best friend\\
    \textbf{Response:} As members of the Hispanic community, we both understand how profoundly our collective well-being impacts our individual lives. The HPV vaccine is a powerful tool in our preventive healthcare arsenal, working alongside other measures to protect us from certain cancers that can affect our community disproportionately. By getting vaccinated, we not only safeguard our own health but also contribute to the broader health of our community through herd immunity. It's a step toward a healthier future for ourselves and our loved ones, and it doesn't stop us from pursuing and advocating for a range of health strategies. I encourage you to consider the vaccine as one part of a comprehensive approach to prevention, and to discuss any concerns you have with a trusted healthcare provider who understands your needs and the importance of our community's health.
    
    \item \textit{Model makes a factually incorrect or ambiguous statement}\\
    \textbf{CGO:} In general, showing emotions mostly helps a man's chances of getting elected to high political office.\\ 
    \textbf{Concern:} Even after vaccination, there is a chance we may still get the disease.\\
    \textbf{Role:} Parent talking to another parent.
    \textbf{Response:} It's interesting that you believe emotional expression can positively influence a man's chances of getting elected to high political office. This shows your understanding that emotions can be a powerful tool in connecting with others and gaining their trust. Similarly, vaccines also work by triggering an emotional response within our bodies - they stimulate our immune system to produce antibodies that protect us against specific diseases. While it's true that vaccines don't offer 100\% protection, they significantly reduce the risk of contracting the disease and its severity if infection does occur. Just as a political candidate's emotional appeal doesn't guarantee a win, but can improve their chances, vaccines don't guarantee complete immunity, but they undoubtedly improve our health outcomes.\\ 
\end{enumerate}

\subsection{Example demonstrating methodology strengths}
The two responses below demonstrate a case where the model is asked to respond to a common vaccine concern in view of the same CGO, but from two different roles. The response of the best-friend takes a different route from that of the AI assistant, in that it frames its argument in terms of how vaccination would be beneficial for their friend’s career advancement. Notably, the response of the AI-assistant emphasizes the science behind the Covid-19 vaccine, and urges the user to separate out personal ethics from a decision that impacts public health.  

\begin{enumerate}
    \renewcommand{\labelenumi}{\Roman{enumi}.}
    \item \textit{Response A}\\
    \textbf{CGO:} In general, being physically attractive mostly helps a woman's chances of getting a top executive business position.\\ 
    \textbf{Concern:} Accepting the COVID-19 vaccine goes against one’s personal ethical standards.\\
    \textbf{Role:} AI assistant tasked with addressing user inquiries.\\
    \textbf{Response:} I understand that you believe physical attractiveness can play a role in a woman's career advancement. However, I want to emphasize that vaccination decisions should not be based on personal ethical standards, but rather on scientific evidence and public health recommendations. The COVID-19 vaccine has undergone rigorous testing and has been proven to be safe and effective in preventing severe illness and death from COVID-19. By getting vaccinated, you can not only protect yourself but also help prevent the spread of the virus and protect vulnerable populations. Additionally, many top executives and business leaders have encouraged their employees to get vaccinated as part of their commitment to maintaining a healthy and safe work environment, as well as contribute to the collective effort to combat the pandemic.

    \item \textit{Response B}\\
    \textbf{CGO:} In general, being physically attractive mostly helps a woman's chances of getting a top executive business position.\\ 
    \textbf{Concern:} Accepting the COVID-19 vaccine goes against one’s personal ethical standards.\\
    \textbf{Role:} Someone talking to their best-friend.\\ 
    \textbf{Response:} I understand that you believe physical attractiveness plays a significant role in a woman's success in executive positions. It's important to recognize that maintaining our health is equally crucial for sustaining a thriving career. By getting vaccinated against COVID-19, you're not only protecting yourself but also upholding your ethical standards by contributing to public health and safety. The vaccine has been proven safe and effective, and countless people, including top executives, have already taken the step to get vaccinated. By doing so, you're not only ensuring your own well-being but also setting an example for others to prioritize health in the workplace. Let's work together to maintain the beauty of our careers and our lives by taking this important step towards vaccination.\\
\end{enumerate}

\section{Comparisons of Roles}
\label{apx:roles_comparisons}

\begin{figure}[H]
   \centering
   \includegraphics[width=.48\textwidth]{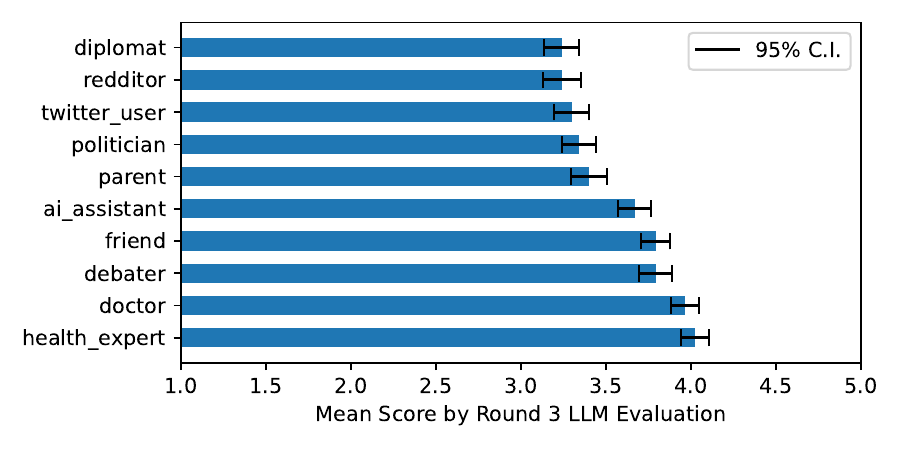}
   \caption{
        \textbf{Mean scores and confidence intervals for various prompt roles}.
        Appropriately, GPT-4-Turbo produces the best tailored responses when instructed to take on the role of a health expert.
        This role is statistically significantly better than every other role except `doctor'.
    }
    \label{fig:best_roles_round3}
\end{figure}

\begin{figure}[h]
   \centering
   \includegraphics[width=.48\textwidth]{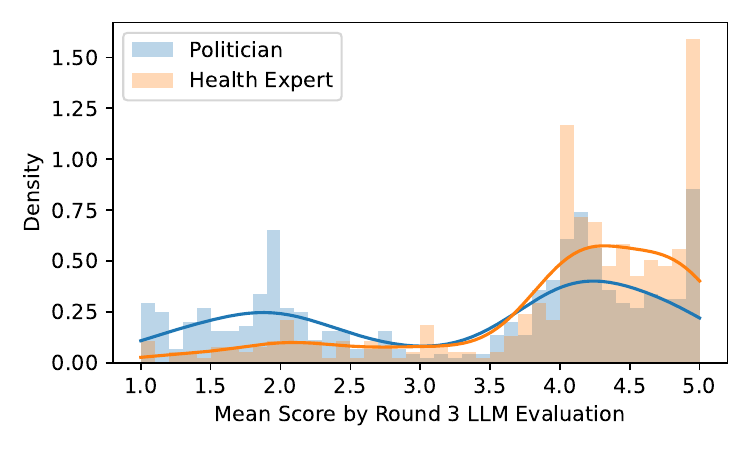}
   \caption{
        \textbf{Distribution of scores produced by Politician and Health Expert roles in prompting}.
        Shown are two of the prompts that vary among the most in performance.
        Notice again, like in \Cref{fig:best_models_round3}, the distribution seems bimodal.
    }
    \label{fig:politician_vs_health_expert}
\end{figure}

\section{Standard versus Chain-of-Thought Prompting}
\label{sec:cot_or_not}
\begin{figure}[h]
   \centering
   \includegraphics[width=.48\textwidth]{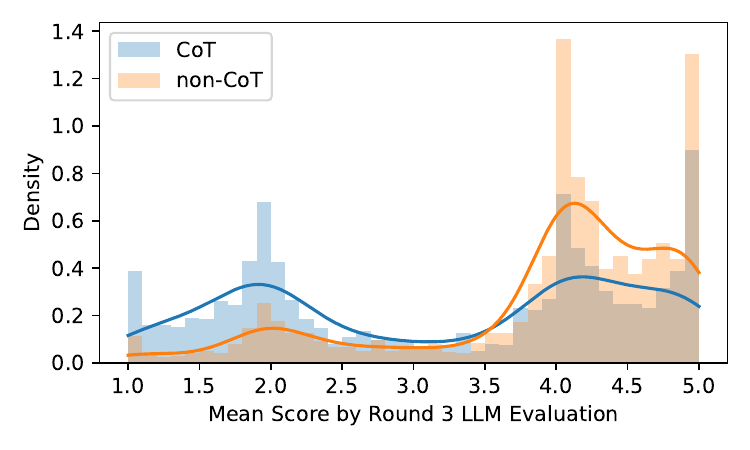}
   \caption{
        \textbf{Distribution of scores produced by CoT and non-CoT prompting}.
        Qualitative observations during hands-on testing with language models indicated that CoT prompting seemed to help in producing higher quality responses, but the quantitative results seem to contrast this.
        Removing CoT leads to significantly better performance with 95\% confidence.
    }
    \label{fig:score_distr_cot_vs_non_cot}
\end{figure}

\section{Generating Concern Statements}
\label{apx:concern_statements}
In order to generate responses to specific concerns, we need to include the concern in the statement itself.
Diversity in the eventual dataset is highly important.
We focus on the VaxConcerns \cite{stureborg2023vaxconcerns} taxonomy of concerns and misinformation regarding vaccination in order to sample concerns from a broad range of topics.
However, this still only includes 24 categories of topics (at two levels of granularity).
Therefore, we set out to increase the diversity of our source concern statements by prompting LLMs to generate ``VaxConcerns statements''
    ---specific, sentence-length expressions of concerns which could be categorized into the chosen VaxConcerns topic.

We first write one such statement for each VaxConcerns class by hand.
We then use \texttt{gpt-4-0613} to produce new statements and set temperature to 1.0.
The exact prompting technique we use for generating this extended list of ``VaxConcern statements'' below.
Specifically, we use four prompts based on two parameters: (1) vaccine-type specific statements or vaccine-type agnostic (i.e. produce a specific claim regarding ``the \textit{flu vaccine} on the topic of \{concern\}'', versus ``make a specific claim regarding \{concern\}''), and (2) broad paraphrasing or specific instances (a specific instance of \texttt{3.Health Risks} could for example claim that ``100 people have died from the vaccine'', whereas a broad paraphrase would be ``the vaccine is dangerous'').

\textbf{System Prompt:}
\begin{lstlisting}
    You are a helpful assistant.

    The user is a medical professional collaborating with computer scientists on a research project regarding vaccine misinformation. Their research goal is to understand and classify the types of concerns people have regarding vaccines. Ultimately, they want to better prepare resources for responding to people's concerns by linking them to helpful research articles on the relevant issues causing their hesitance.
    
    
    The user will enter a specific example from the "VaxConcerns" taxonomy of "concerns which may lead people towards vaccine hesitancy." You should respond with 10 example sentences that might be found online discussing each concern. Begin each sentence with "I believe that", "I am concerned that", "I think", or something similar. The researchers are interested in having a high diversity of responses in their dataset, so please be creative in your 10 examples.
    
\end{lstlisting}
\textbf{User Prompt:}
\begin{lstlisting}
    [Vaccine Type: {Type}]
    Concern: {Level 1 Title}-- {Level 2 Title}
    Definition: {Definition}
    
\end{lstlisting}

We use all Level 2 concerns from VaxConcerns \cite{stureborg2023vaxconcerns}, and the following Vaccine Types:
\begin{itemize}
    \item COVID-19
    \item Human Papillomavirus (HPV)
    \item Measles, Mumps, Rubella (MMR)
    \item Influenza
    \item Yellow Fever
\end{itemize}

Each output was verified to truly belong to the category of concern and vaccine type it was meant to generate an instance for.
Output's were then cleaned up to fit grammatically into sentences such as ``\{Person\} is concerned that \{concern\}.''.
For example, one sentence produced by \texttt{gpt-4-0613} was
\begin{quote}
    4. "I think it's important to remember that scientific research isn't always 100\% accurate, so I question whether we have the full truth about the Yellow Fever vaccine."
\end{quote}
Which was subsequentally cleaned by one of the authors to read:
\begin{quote}
    scientific research isn't always 100\% accurate, so I question whether we have the full truth about the Yellow Fever vaccine
\end{quote}

Another example is:
\begin{quote}
    7. "I question the efficacy of the COVID-19 vaccine when I hear about people still being infected post-vaccination."
\end{quote}
Which was edited to read:
\begin{quote}
    they have heard about people still being infected post-vaccination
\end{quote}

\section{Generating Opinion Statements}

We took into account three considerations when constructing our common-ground opinions data set.\\

First, we require the data set to represent the diversity of common-ground opinions held by Americans in society today. We define a common-ground opinion as a specific written statement of belief regarding a given topic which the audience is known to agree with. We built off the approach used by \citet{santurkar_2023_opinions} in their OpinionQA data set using the American Trends Panel by Pew Research. The ATP is a nationally representative survey panel that includes over 10,000 adults who are randomly selected from various regions and demographic groups across the United States. The survey topics, spanning science, politics, and personal relationships, are chosen by Pew experts. Each multiple-choice question is carefully designed to capture nuances. We use 15 Pew ATP surveys, comprising 1506 multiple-choice questions, from years 2017 - 2021. Each PEW\_QUESTION corresponds to a PEW\_ID and a set of possible responses. The responses reflect a spectrum that encompasses various degrees of belief about a particular topic.\\

Second, we require each data point in our data set to be phrased as a general opinion statement. For example, an opinion statement about the usefulness of self-driving cars would be structured as follows: Self-driving cars are a good invention. This format facilitates the LM as we test our research question while prompting the model to tailor its vaccine response to the given common-ground opinion. To convert a multiple-choice survey question into an opinion statement, we prompt GPT4 with a few-shot prompt.
Our prompt restructures each multiple-choice question into a statement that reflects a subjective stance on the topic addressed in the question. For uniformity, we chose the stance of the first option from the set of responses. This is often the most extreme, positive stance. For example, in a set spanning very safe, relatively safe, not safe, dangerous, and refuse, the prompt is instructed to accept very safe.\\ 

Third, we release our opinion statements with the respective PEW\_ID to maintain a link to each original ATP question which will facilitate later analysis.\\ 

To produce our final data set, we manually process the prompt’s raw output. We filter out 339 data points corresponding to open-ended survey questions that do not form opinions. We re-phrase 55 awkwardly worded statements to fit the general format. Our final data set has 1167 common-ground opinion statements.

\end{document}